\documentclass[runningheads]{llncs}
\pdfoutput=1
\usepackage[pdftex]{graphicx}
\usepackage{amsmath}
\usepackage{amssymb}
\usepackage{mathtools}
\usepackage{epstopdf}
\usepackage{bm}
\usepackage{mathtools}
\usepackage{multirow}
\usepackage{todonotes}

\DeclareMathOperator*{\argmax}{arg\,max}

\newcommand{\never}{\textsc{NeVer}}
\newcommand{\nevertwo}{\textsc{NeVer 2.0}}
\newcommand{\shark}{\textsc{shark}}
\newcommand{\hysat}{\textsc{HySAT}}
\newcommand{\pytorch}{\textsc{PyTorch}}
\newcommand{\tensorflow}{\textsc{TensorFlow}}

\newcommand{\conc}[1]{\textsf{#1}}
\newcommand{\absc}[1]{\textsf{\textit{#1}}}

\begin{document}
\title{\nevertwo{}: Learning, Verification and Repair\\
  of Deep Neural Networks}
\titlerunning{NeVer 2.0}
%
\author{Dario Guidotti\inst{1}  \and
Luca Pulina\inst{2} \and
Armando Tacchella\inst{1} }
\authorrunning{Guidotti et al.}
%
\institute{University of Genoa, Italy\and
University of Sassari, Italy\\
\email{dario.guidotti@edu.unige.it,\\ lpulina@uniss.it, armando.tacchella@unige.it}}
\maketitle              
\begin{abstract}

In this work we present an early prototype of \nevertwo{},
a new system for automated synthesis and analysis of deep neural
networks. \nevertwo{} borrows its design philosophy from \never{}, the
first package that integrated learning, automated verification and
repair of (shallow) neural networks in a single tool. The goal of
\nevertwo{} is to provide a similar integration for deep networks 
by leveraging a selection of state-of-the-art learning frameworks
and integrating them with verification algorithms to ease the
scalability challenge and make repair of faulty networks possible.    

\end{abstract}
%
%
%
\section{Introduction}
\label{sec:intro}
Adoption and successful application of deep neural networks (DNNs)
in various domains have made them one of the most popular
machine-learned models to date --- see,
e.g.,~\cite{DBLP:conf/cvpr/TaigmanYRW14} on image
classification,~\cite{DBLP:journals/taslp/YuHMCS12} on speech
recognition, and~\cite{DBLP:journals/nature/LeCunBH15} for the general
principles and a catalog of success stories.
Despite the impressive progress that the learning community has made 
with the adoption of DNNs, it is well known that their application in
safety- or security-sensitive contexts is not yet hassle-free. 
From their well-known sensitivity to \textit{adversarial
perturbations}~\cite{DBLP:journals/corr/SzegedyZSBEGF13,DBLP:journals/corr/GoodfellowSS14},
i.e., minimal changes to correctly classified input data that 
cause a network to respond in unexpected and incorrect ways, 
to other less-investigated, but possibly significant properties ---
 see, e.g.,~\cite{DBLP:journals/corr/abs-1805-09938} for a catalog ---
 the need for tools to analyze and possibly repair DNNs is strong.

As witnessed by an extensive survey~\cite{huang2018safety}
of more than 200 recent papers, the response from the scientific
community has been equally strong. 
As a result, many algorithms have been proposed for verification of
neural networks and tools implementing them have been made
available. Some examples of well-known and fairly mature verification
tools are Marabou~\cite{DBLP:conf/cav/KatzHIJLLSTWZDK19}, an SMT-based
tool that answers queries regarding the properties of a DNN by
transforming the queries into constraint satisfiability problems;
ERAN~\cite{DBLP:conf/iclr/SinghGPV19}, a robustness analyzer based on
abstract interpretation and MIPVerify~\cite{DBLP:conf/iclr/TjengXT19},
another robustness analyzer based on mixed integer programming
(MIP). Other widely-known verification tools are
Neurify~\cite{DBLP:conf/nips/WangPWYJ18}, a robustness analyzer based
on symbolic interval analysis and linear relaxation,
NNV~\cite{DBLP:journals/corr/abs-2004-05519}, a tool implementing
different methods for reachability analysis,
Sherlock~\cite{DBLP:conf/hybrid/DuttaCJST19}, an output range analysis
tool and NSVerify~\cite{DBLP:conf/kr/AkitundeLMP18}, also for
reachability analysis. A number of verification methodologies ---
without a corresponding tool --- is also available
like~\cite{DBLP:journals/corr/abs-1807-03571}, a game based 
methodology for evaluating pointwise robustness of neural networks in
safety-critical applications. Most of the above-mentioned tools and
methodologies work only for feedforward fully-connected neural
networks with ReLU activation functions, with some of them
featuring verification algorithms for convolutional neural networks 
with different kinds of activation function. To the best of our
knowledge, current state-of-the-art tools are restricted to
verification/analysis tasks, in some cases they are limited to
specific network architectures and they might prove difficult to use
for the non-initiated.

In this work we present an early prototype of \nevertwo{},
a new system that aims to bridge the gap between learning and
verification of DNNs and solve some of the above
mentioned issues. \nevertwo{} borrows its design
philosophy from \never{}~\cite{pulina2011n}, the first tool for 
automated learning, analysis and repair of neural networks. \never{}
was designed to deal 
with multilayer perceptrons (MLPs) and its core  was an
abstraction-refinement mechanism described
in~\cite{DBLP:conf/cav/PulinaT10,DBLP:journals/aicom/PulinaT12}.  
As a system, one peculiar aspect of \never{} was that it included
learning capabilities through the \shark~\cite{igel08} library.
Concerning the verification part, \never{} could utilize any
solver integrating Boolean reasoning and linear arithmetic constraint
solving --- \hysat~\cite{franzle2007efficient} at the time.
A further peculiarity of the approach was that \never{}
could leverage abstract counterexamples to (try to) repair the
MLP, i.e., retrain it to eliminate the causes of misbehaviour. 

Our goal for \nevertwo{} is to provide the same features of \never{},
but in an updated package that has the following features:
\begin{itemize}
\item Loading of datasets, trained and untrained models provided in a
variety of formats; currently \nevertwo{} supports directly popular
datasets, e.g., MNIST~\cite{lecun1998gradient} and Fashion
MNIST~\cite{DBLP:journals/corr/abs-1708-07747}, but support
for further datasets can be added through a common interface;
models (either trained or not) can be supplied to \nevertwo{} using
ONNX\footnote{\url{https://onnx.ai/}} and \pytorch{}\footnote{\url{https://pytorch.org/}}~\cite{DBLP:conf/nips/PaszkeGMLBCKLGA19}
formats --- \tensorflow{}\footnote{\url{https://www.tensorflow.org/}}~\cite{DBLP:journals/corr/AbadiBCCDDDGIIK16}
support is under development.
\item Training of DNNs through state-of-the-art frameworks;
currently \nevertwo{} is based on \pytorch{},
but further extensions are planned to handle different kinds of learning
models (e.g., kernel-based machines) that are not handled natively
by \pytorch{}, or to leverage specific capabilities of other learning
frameworks. 
\item Manipulation of DNNs including, but not limited to,
pruning~\cite{DBLP:conf/icnn/SietsmaD88}, quantization~\cite{DBLP:journals/tnn/XieJ92},
and transfer learning~\cite{torrey2010transfer};
currently \nevertwo{} builds on \pytorch{} to manipulate DNNs, and
only two mainstream pruning techniques are 
implemented, namely network slimming~\cite{DBLP:conf/icnn/SietsmaD88}
and weight pruning~\cite{DBLP:conf/nips/CunDS89}. 
\item Verification of DNNs: currently, \nevertwo{} leverages external
tools as backends to provide verification capabilities; connectors to
Marabou~\cite{DBLP:conf/cav/KatzHIJLLSTWZDK19},
ERAN~\cite{DBLP:conf/iclr/SinghGPV19}  
and MIPVerify~\cite{DBLP:conf/iclr/TjengXT19} are currently
implemented; we plan to add abstraction-refinement algorithms
that improve on and extend those available in \never{}, but their
development is still underway.
\item Repair of DNNs should enable the results of verification to
improve on the results of learning; currently, \nevertwo{} features
the same mechanism of \never{}, i.e., it relies on the capability of
the embedded learning algorithms to exploit counterexamples 
and retrain the network in a better way --- a sort of adversarial
training guided by verification; we expect to reach tighter integration
of verification and learning once our custom verification algorithms
are implemented.
\end{itemize}

The version of \nevertwo{} corresponding to this work is available online~\cite{never2}
under the Commons Clause (GNU GPL v3.0) license.\\
The rest of the paper is structured as follows. In
Section~\ref{sec:ml} we introduce some basic notations and definition 
to be used through the paper. In Section~\ref{sec:architecture} we
describe the architecture and the current implementation
of \nevertwo{}. In Section~\ref{sec:results} we show some early
results obtained with \nevertwo{} prototype using MNIST datasets
to learn and verify fully-connected ReLU networks. We conclude the
paper with and our future research agenda in 
Section~\ref{sec:extensions}.

\section{Preliminaries}
\label{sec:ml}
%
A \emph{neural network} is a system of interconnected computing
units called \emph{neurons}.  
In \emph{fully connected feed-forward} networks, neurons are arranged in
disjoint layers, with each layer being fully connected only with
the next one, and without connection between neurons in the same
layer. Given a feed-forward neural network $N$ with $n$ layers, we
denote the $i$-th layer of $N$ as $\mathbf{h}^{(i)}$. We call a layer
without incoming connections \emph{input} layer $\mathbf{h}^{(1)}$,
a layer without outgoing connections \emph{output} layer
$\mathbf{h}^{(n)}$, while all other layers are referred to as 
\emph{hidden} layers . 
Each hidden layer performs specific transformations on the inputs
it receives. In this work we consider hidden layers that
make use of \emph{linear}  and \emph{batch normalization}
modules. Given an input vector $\mathbf{x} $, a linear module computes
a linear combination of its values as follows:   
\begin{align}
\begin{split}
&\mathbf{L}^{(i)} = \mathbf{W}^{(i)} \cdot \mathbf{x} + \mathbf{b}^{(i)}
\label{eq:dnn-ext}
\end{split}
\end{align}
where $\mathbf{W}^{(i)} $ is the matrix of weights and 
$\mathbf{b}^{(i)} $ is the vector of
the biases associated with the linear module in the $i$-th layer and $\mathbf{L}^{(i)} $ is the corresponding output. Entries of both $\mathbf{W}^{(i)}$ and $\mathbf{b}^{(i)}$ are learned parameters.
In our target architectures, each linear module is followed by a batch normalization module. This is done
to address the so-called \emph{internal covariate shift} problem,
i.e., the change of the distribution of each layer's input during
training~\cite{DBLP:conf/icml/IoffeS15}.  
The mathematical formulation of batch normalization layers can be expressed as 
\begin{equation}
\mathbf{BN}^{(i)}  = \frac{\bm{\gamma}^{(i)}}{\sqrt[2]{\bm{\sigma}^{(i)} + \epsilon^{(i)}}} \odot (\mathbf{L}^{(i)} - \bm{\mu}^{(i)}) + \bm{\beta}^{(i)}
\label{eq:batch-norm}
\end{equation}
All the operators in this equation are element-wise operators: in 
particular $\odot$ and the fractional symbol represent respectively
the Hadamard 
product and division. 
$\mathbf{BN}^{(i)}$ and $\mathbf{L}^{(i)}$ are the 
output and the input vectors of the module, respectively.
$\bm{\gamma}^{(i)}$, $\bm{\mu}^{(i)}$, $\bm{\sigma}^{(i)}$,
$\bm{\beta}^{(i)}$ are vectors, whereas $\epsilon^{(i)}$
is a scalar value. These are learned parameters of the batch normalization
layer. In particular $\bm{\mu}^{(i)}$ and  $\bm{\sigma}^{(i)}$ are
the estimated mean and variance of the inputs computed during training.\\
Finally, the output of hidden layer $i$ is computed as $\mathbf{h}^{(i)} = \Phi^{(i)} (\mathbf{BN}^{(i)})$,
where $\Phi^{(i)}$ is the \emph{activation function} associated to
the neurons in the layer. We consider only networks utilizing
\emph{Rectified Linear Unit (ReLU)} activation functions, i.e.,  $\Phi^{(i)} = \max(0, \mathbf{BN}^{(i)})$.
Given an input vector $\mathbf{x}$, the network $N$ computes an output vector $\mathbf{y}$ by means of the
following computations
\begin{align}
\begin{split}
& \mathbf{h}^{(1)} = \mathbf{x}\\
& \mathbf{h}^{(i)} = \Phi^{(i)} (\mathbf{BN}^{(i)}(\mathbf{L}^{(i)} (\mathbf{h}^{(i-1)}))) \quad i=2,\ldots,n-1\\
& \mathbf{y} = \mathbf{h}^{(n)} = \mathbf{L}(\mathbf{h}^{(n-1)})
\label{eq:dnn}
\end{split}
\end{align}

A neural network can be considered as a non-linear function
$f_\textbf{w}: \mathcal{X} \rightarrow \mathcal{Y}$, where
$\mathcal{X}$ is the input space of the network, $\mathcal{Y}$ is
the output space and $\mathbf{w}$ is the vector representing the weights of all the connections. We consider neural network applied to classification 
of $d$-dimensional vectors of real numbers, i.e., 
$\mathcal{X} \subseteq \mathbb{R}^d$ and $\mathcal{Y} \subseteq \mathbb{R}^m$, where $d$ is the dimension of the input vector and $m$ is the dimension of the output vector and thus also the number of possible classes of interest.
We assume that given an input sample $\mathbf{x}$ the output vector
$f_\mathbf{w}(\mathbf{x})$ contains the likelihood that $\mathbf{x}$
belongs to one of the $m$ classes. The specific class can be computed
as
$$
\argmax\limits_{c \in \{1, \ldots, m\}} (f_{\mathbf{w}}(\mathbf{x}))_c
$$
where $(f_{\mathbf{w}}(\mathbf{x}))_c$ denotes the $c$-th element of
$f_\mathbf{w}$.
Training of (deep) neural networks poses substantial computational challenges
since for state-of-the-art models the size of $\mathbf{w}$ can be in the order of millions.
As in any machine learning task, training must select weights
to maximize the likelihood that the network responds correctly,
i.e., if the input $\mathbf{x}$ is of class $k$, the chance of
\emph{misclassification} should be as small as possible, where
misclassification occurs whenever the following holds  
$$
\argmax\limits_{c \in \{1, \ldots, m\}} (f_{\mathbf{w}}(\mathbf{x}))_c \neq k
$$
Training can be achieved through minimization of some kind of
\emph{loss function} whose value is low when the chance of
misclassification is also low. While there are many different kinds of
loss functions, in general they are structured in the following way:     
\begin{equation}
J(\mathbf{w}) = \frac{1}{n} \sum_{k=0}^n Err(y_k, \argmax\limits_{c \in \{1, \ldots, m\}} (f_{\mathbf{w}}(\mathbf{x}_k))_c) + \lambda \cdot Reg(\mathbf{w})
\end{equation}
where $n$ is the number of training pairs $(\mathbf{x}_k, y_k)$,
$y_k$ is the correct class label of $\mathbf{x}_k$, 
$Err$ represents the loss caused by misclassification, 
$Reg$ is a \emph{regularization} function, and $\lambda$ is the
parameter controlling the effect of $Reg$ on $J$.
The regularization function is needed to avoid \emph{overfitting},
i.e., the high variance of the training results with respect to the
training data. The regularization function usually penalizes models with high complexity by smoothing out sharp variations induced in the trained network by the $Err$ function. A common
regularization function is, for example, the L2 norm: 
\begin{equation}
Reg(\mathbf{w}) = \frac{1}{2n} ||\mathbf{w}||_2
\end{equation}
\section{System architecture and implementation}
\label{sec:architecture}
\begin{figure}
\centering
\scalebox{0.4}{\includegraphics{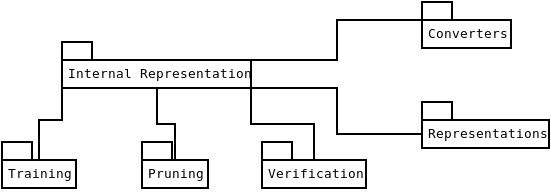}}
\caption{\label{fig:packages} High level UML diagram of \nevertwo{}.}

\end{figure}

\nevertwo{} is conceived as a modular API to manage DNNs, from
training to verification and repair. In Figure~\ref{fig:packages} we
present an overview of the architecture divided in six main
packages. The main elements we consider in this work are training,
pruning and verification: these packages are organized mostly around
\emph{Strategy} patterns that define general interfaces to perform
network operations, and specialized subclasses that actually
support those operations. Additionally, to have full control of the
internal model and to separate the main elements from implementation
details, we designed our own network representation structured as a 
graph whose nodes correspond to disjoint layers. To leverage the
capabilities of current learning frameworks, we designed a set of
conversion strategies to/from our internal representation and the
representations whereon learning frameworks are based. The aims and
the internal structures of the packages shown in
Figure~\ref{fig:packages} are described in detail in the remainder of
this Section.   

\subsection{Internal Representation}
\label{subsec:internal-representation}

\begin{figure}
\centering
\scalebox{0.33}{\includegraphics{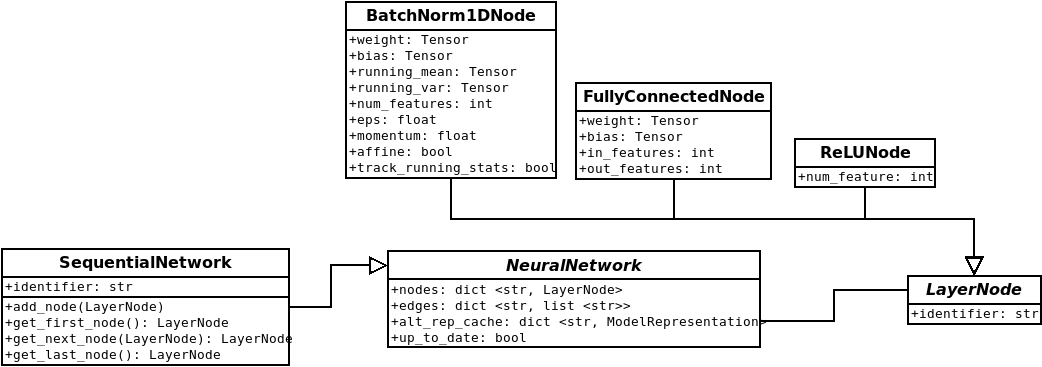}}
\caption{\label{fig:network}UML diagram of internal network
  representation.}
\end{figure}

The classes supporting the internal representation are shown
in Figure~\ref{fig:network}. There are two abstract base classes,
namely \absc{NeuralNetwork} and \absc{LayerNode}. Conceptually,
\absc{NeuralNetwork} is a container of \absc{LayerNode} objects
orgnaized inside \absc{NeuralNetwork} as a graph. A list of 
\absc{ModelRepresentation} objects is kept for internal usage --- see
subsection~\ref{subsec:converters} for details. In the current
prototype the only concrete subclass of \absc{NeuralNetwork} is
\conc{SequentialNetwork} which represents networks whose
corresponding graph is a list, i.e., each layer is connected
only to the next one. More complex topologies for concrete
architectures can be implemented, should the need arise. The concrete
subclasses of \absc{LayerNode} are the building blocks that we
currently support: \conc{BatchNorm1DNode}, 
\conc{FullyConnectedNode} and \conc{ReLUNode}, i.e., 
batch normalization layers, fully connected layers and
ReLU layers, respectively. These building blocks are sufficient to encode
the DNNs that we introduced in Section~\ref{sec:ml}. It should be noted
that our representation is not an ``executable'' representation, i.e., 
it does not provide the capability to compute the output of a DNN given the 
input, therefore our nodes contain only enough information
to create the corresponding executable representations in different
learning frameworks and/or support the encoding for verification
purposes.  The class \absc{Tensor} is our utility class 
for tensorial data. Currently it is simply an alias for
the \texttt{ndarray} class in \texttt{numpy}, but we have added it as a
wrapper to isolate \nevertwo{} classes from implementation details.

\subsection{Converters and Representations}
\label{subsec:converters}

\begin{figure}
\centering
\scalebox{0.28}{\includegraphics{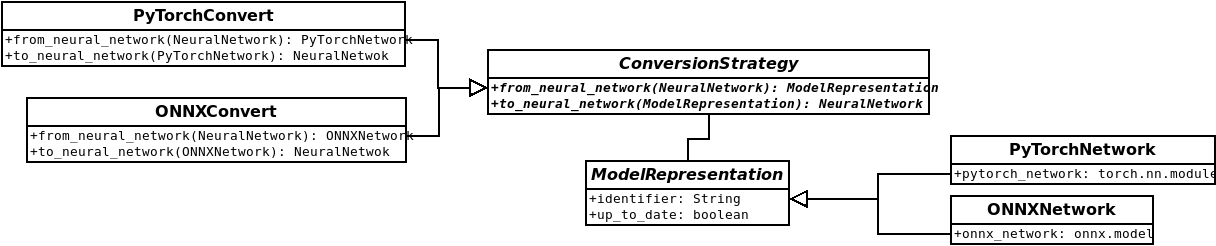}}
\caption{\label{fig:conv}UML diagram of the classes related to the representations and converters of the
different learning frameworks.}
\end{figure}

The design of a model representation to generalize those used in 
different learning frameworks is based on the \emph{Adapter} design
pattern, as shown in Figure~\ref{fig:conv}. We have defined the
abstract class \absc{ModelRepresentation}.  which is then specialized
by \conc{PyTorchNetwork} and \conc{ONNXNetwork} to encode \pytorch{} and
ONNX models, respectively. The concrete subclasses wrap the actual
network model in the corresponding learning framework or
interchange format, as in the case of ONNX. Conversion between our internal 
representation and the concrete subclasses of
\absc{ModelRepresentation} are provided by subclasses of
\absc{ConversionStrategy} --- we may consider this also as a \emph{Builder}
pattern implementation. \absc{ConversionStrategy} defines an interface
with two functions: one for converting from our internal representation
to a specific model representation, and the other for performing
the inverse task. The concrete subclasses of \absc{ConversionStrategy}
implement the functions for the corresponding concrete subclasses of
\absc{ModelRepresentation}. As new type of learning frameworks/model
are added to \nevertwo, new concrete subclasses of
\absc{ModelRepresentation} should be added to support conversion.

\subsection{Training}
\label{subsec:training}

\begin{figure}
\centering
\scalebox{0.29}{\includegraphics{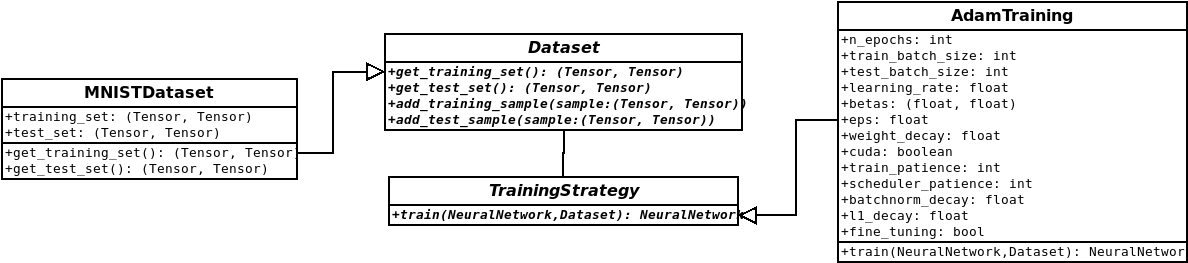}}
\caption{\label{fig:training}UML diagram of the classes related to the training strategies.}
\end{figure}

In Figure~\ref{fig:training} we show the internal design of the
Training package whose main element is the abstract class
\absc{TrainingStrategy}. The current abstraction of a training
strategy features a single function which requires a
\absc{NeuralNetwork} and a \absc{Dataset} and returns a (trained)
\absc{NeuralNetwork}. The concrete subclasses of
\absc{TrainingStrategy} provide the actual training procedures.
Currently, we have designed and implemented 
a single training procedure based on the Adam
optimizer~\cite{DBLP:journals/corr/KingmaB14} and adapted to the 
concrete pruning procedures we have implemented. Our implementation
requires a \pytorch{} representation to train the network, but this is
handled transparently by \nevertwo{} architecture. The class
\absc{Dataset} is meant to represent a generic dataset. As such
it features four functions: one for loading the training set --- the
set of data considered to train the network --- one for loading
the test set -- the set of data considered to assess the accuracy of
the network and two for adding a data sample to the training set and to
the test set respectively. The actual datasets are represented by concrete
subclasses of \absc{Dataset}. Currently, we have implemented the
corresponding concrete class for the MNIST dataset \conc{MNISTDataset}
and for the FMNIST dataset \conc{FMNISTDataset}.

\subsection{Pruning}
\label{subsec:pruning}

\begin{figure}
\centering
\scalebox{0.3}{\includegraphics{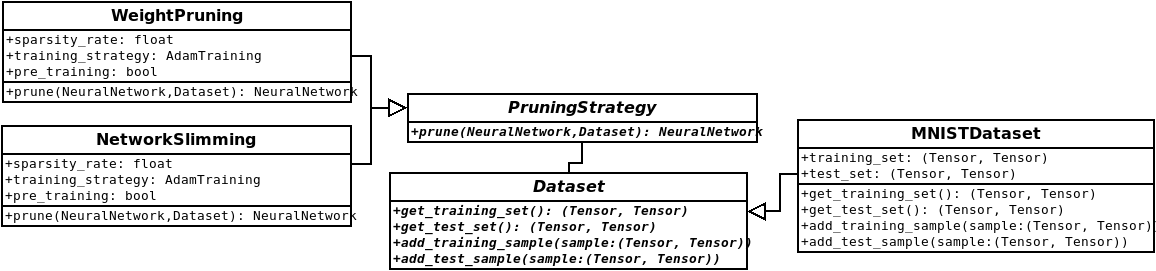}}
\caption{\label{fig:pruning}UML diagram of the classes related to the pruning strategies.}
\end{figure}

As mentioned in our paper~\cite{guido2020ecai}, we believe that
pruning can be one of the keys to ease the verification of DNNs,
therefore  we decided to include abstractions and concrete classes to 
support pruning in the current realization of \nevertwo{}.
In Figure~\ref{fig:pruning} we show the architecture, where the
abstract class \absc{PruningStrategy} is meant to represent a generic
pruning methodology, and consists of a single function which requires
a \absc{NeuralNetwork} and a \absc{Dataset} and returns the pruned
\absc{NeuralNetwork}. Concrete subclasses implement the actual pruning
procedures: currently we have designed and implemented two concrete
strategies, namely \conc{WeightPruning} and \conc{NetworkSlimming} ---
both based on \pytorch{} representations. In particular, the former
strategy selects all the weights which are smaller than
a certain threshold and sets them to 0. The latter strategy leverages
the weights of the batch normalization layers to identify
low-importance neurons and remove them from the network --- more
details can be found in~\cite{DBLP:journals/corr/HanPTD15} 
and~\cite{DBLP:conf/iccv/LiuLSHYZ17}. The distinctive parameters
of the pruning strategies are provided as attributes in
the concrete classes. In particular, if pre-training and/or
fine-tuning are required for the pruning procedure then a suitable 
training strategy must be provided to the pruning strategy as an attribute.

\subsection{Verification}
\label{subsec:verification}

\begin{figure}
\centering
\scalebox{0.3}{\includegraphics{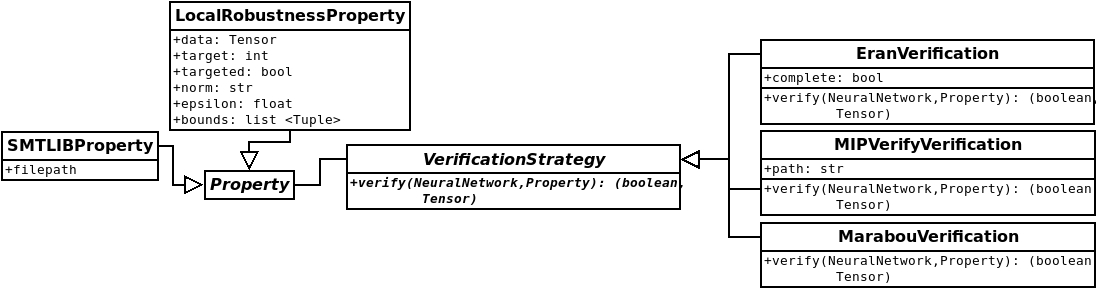}}
\caption{\label{fig:verification}UML diagram of the classes related to the verification strategies.}
\end{figure}
As shown in Figure~\ref{fig:verification}, we have designed the
abstract class \absc{VerificationStrategy} to represent a generic
verification methodology. This abstract class defines an interface
consisting of a single function which requires a \absc{NeuralNetwork}
and a \absc{Property} and returns a Boolean value depending on whether
the property is verified or not and a counterxample (if available).
The abstract class \absc{Property} 
represents a generic property that should be verified. Currently we have
two concrete classes: \conc{SMTLIBProperty} and \conc{LocalRobustnessProperty}.
\conc{SMTLIBProperty} represent a generic property which \nevertwo{}
reads from a file formatted according to SMTLIB~\footnote{http://smtlib.cs.uiowa.edu/} syntax~\cite{BarST-SMT-10}. \conc{LocalRobustnessProperty} is a ``pre-cooked'' property encoding the search of an adversarial example corresponding to a specific data sample.   
The concrete subclasses of \absc{VerificationStrategy} that we have
implemented so far are \conc{EranVerification},
\conc{MarabouVerification} and \conc{MIPVerifyVerification} which
leverage, respectively, ERAN~\cite{DBLP:conf/iclr/SinghGPV19},
Marabou~\cite{DBLP:conf/cav/KatzHIJLLSTWZDK19} and MIPVerify to verify
the property.

\section{Preliminary experimental analysis}
\label{sec:results}
\begin{table}
  \begin{center}
\scalebox{0.7}{%
\begin{tabular}{|c|c|c|c|c|c|c|c|c|}
\hline
\multicolumn{6}{|c|}{\textbf{MNIST}} & \multicolumn{3}{|c|}{\textbf{FMNIST}}\\
\hline
\textbf{Base} & \textbf{Param} & \textbf{Network} & \textbf{Marabou} & \textbf{MIPVerify} & \textbf{ERAN} & \textbf{Marabou} & \textbf{MIPVerify} & \textbf{ERAN}\\
\hline
\multirow{7}{4em}{\emph{NET1}} & \multirow{1}{4em}{} & Baseline & $0$ & $0$ & $0$ & $0$ & $0$ & $1$\\
\cline{2-9}
& \multirow{3}{4em}{\textbf{SET1}} & Sparse & $0*$ & $15*$ & $20$ & $0*$ & $20$ & $20$\\
& & WP & $8$ & $5*$ & $14$ & $2$ & $5$ & $19$\\
& & NS & $18$ & $16*$ & $20$ & $19$ & $20$ & $20$\\
\cline{2-9}
& \multirow{3}{4em}{\textbf{SET2}} & Sparse & $0$ & $0$ & $1$ & $0$ & $0$ & $0$\\
& & WP & $0$ & $0$ & $7$ & $0$ & $0$ & $5$\\
& & NS & $5$ & $15$ & $17$ & $6$ & $4$ & $20$\\
\cline{2-9}
& \multirow{3}{4em}{\textbf{SET3}} & Sparse & $0$ & $0$ & $1$ & $0$ & $0$ & $0$\\
& & WP & $0$ & $0$ & $1$ & $3$ & $0$ & $3$\\
& & NS & $6$ & $0$ & $7$ & $0$ & $0$ & $4$\\
\hline
\multirow{7}{4em}{\emph{NET2}} & \multirow{1}{4em}{} & Baseline & $0$ & $0$ & $0$ & $0$ & $0$ & $1$\\
\cline{2-9}
& \multirow{3}{4em}{\textbf{SET1}} & Sparse & $0*$ & $17*$ & $20$ & $0*$ & $19*$ & $20$\\
& & WP & $9$ & $0$ & $0$ & $0$ & $0$ & $0$\\
& & NS & $18$ & $14*$ & $19$ & $0*$ & $17$ & $20$\\
\cline{2-9}
& \multirow{3}{4em}{\textbf{SET2}} & Sparse & $0*$ & $0$ & $0$ & $0*$ & $0$ & $0$\\
& & WP & $0$ & $0$ & $0$ & $0$ & $0$ & $0$\\
& & NS & $0$ & $0$ & $1$ & $0$ & $0$ & $0$\\
\cline{2-9}
& \multirow{3}{4em}{\textbf{SET3}} & Sparse & $0$ & $0$ & $0$ & $0$ & $0$ & $0$\\
& & WP & $0$ & $0$ & $0$ & $0$ & $0$ & $0$\\
& & NS & $1$ & $0$ & $0$ & $0$ & $0$ & $0$\\
\hline
\end{tabular}
}
\end{center}
\caption{Results --- originally reported in~\cite{guido2020ecai} ---
  obtained by running \nevertwo{} with Marabou, ERAN and
  MIPVerify. The values reported represent the number of problems
  which were solved successfully within the timeout of 600 CPU
  seconds.  The column \textbf{Base} represent the base architecture,
  \textbf{Param} represent the set of parameters used for increasing
  magnitude of pruning and \textbf{Network} represent the kind of
  network considered.  \textbf{Marabou}, \textbf{MIPVerify}, and
  \textbf{ERAN} represent the number of problems solved by Marabou,
  MIPVerify and ERAN, respectively.}
\label{tab:ver-result}
\end{table}

We test the current capabilities of \nevertwo{} by replicating the
setup of the experiment reported in~\cite{guido2020ecai}. In this
experiment we analyze how the integration of pruning and verification
can ease analysis of DNNs --- currently, a distinctive capability
that \nevertwo{} offers. 
We test two different network architectures
and two different pruning methods considering all three verification
backends available in \nevertwo{}.
The DNNs we consider are both fully connected networks with
three hidden layers: one with 64, 32, 16 hidden neurons, and the other
with 128, 64, 32 hidden neurons. In both networks the activation
function is the ReLU. We experimented with weight pruning (based  
on~\cite{DBLP:journals/corr/HanPTD15}) and network slimming (based
on~\cite{DBLP:conf/iccv/LiuLSHYZ17}).  To analyze the performances of
the different pruning methods we test them with three different
sets of pruning parameters which correspond to increasing magnitudes
of pruning. The results of our experiment are summarized in Table
\ref{tab:ver-result}. We consider three versions of each DNN:
the version before pruning (baseline), the one obtained after a
specialized training for network slimming (sparse), the version
after the application of weight pruning (WP) and the one
obtained after network slimming (NS). The results of our experiment 
prove that \nevertwo{} --- albeit still at the prototypical stage --
is ready to verify networks of some practical interest, and its
combination of pruning and verification may offer some advantage over
the straight usage of its backends.

\section{Planned extensions}
\label{sec:extensions}
\nevertwo{} is an ongoing project and we have already planned various
extensions. First, we aim to increase the variety of networks that can
be represented by adding more concrete subclasses to
\emph{LayerNode}. In particular, we expect to be able to design and
implement convolutional layers, the related batch normalization
layers, different kinds of pooling layers and different kinds of
activation functions. With these extensions, that should be
matched by corresponding training, pruning and verification
enhancements, \nevertwo{} should be able to represent all the 
main kinds of DNNs which current state-of-the-art verification
methodologies can deal with. 

The second addition that we have already planned, relates to
the addition of converters to/from other major learning frameworks,
starting from \tensorflow{}. This addition should include also the
capability of visualizing and modifying the network architecture
through a graphical user interface, in the hope that \nevertwo{}
becomes more easily accessible also to the non-initiated.

Further additions that we wish to add include quantization, which we
believe would create interesting synergies with pruning, and repair,
i.e,  the capability to modify a neural network to make it compliant
to the property of interest. In particular, besides basic form of
repair that are already supported by \nevertwo{}, i.e.,
verification-based adversarial learning, we expect to provide tighter
integration between learning and verification.



\newpage
\bibliographystyle{splncs04} \bibliography{biblio_ecai}

\end{document}